\documentclass[runningheads]{llncs}
\usepackage[T1]{fontenc}
\usepackage[symbol]{footmisc}
\usepackage{graphicx, floatrow}
\usepackage{float}
\floatstyle{plaintop}
\restylefloat{table}
\usepackage{booktabs} 
\usepackage{amsmath,url,graphicx,amssymb}
\usepackage{amsfonts}
\usepackage{multirow}
\usepackage{algorithm}
\usepackage{algpseudocode}
\usepackage{pifont}
\usepackage{color}
\usepackage{xspace}
\usepackage{enumitem}
\usepackage{subcaption}
\usepackage{colortbl}
\usepackage{xcolor}
\usepackage{bm}
\usepackage{caption} 
\usepackage{bbm}
\usepackage{todonotes}
\usepackage[
  separate-uncertainty = true,
  multi-part-units = repeat
]{siunitx}

\newcommand{\mylistbegin}{
  \begin{list}{$\bullet$}
   {
     \setlength{\itemsep}{-2pt}
     \setlength{\leftmargin}{1em}
     \setlength{\labelwidth}{1em}
     \setlength{\labelsep}{0.5em} } }
\newcommand{\mylistend}{
   \end{list}  }

\newcommand{\ie}{\textit{i.e.}}

\definecolor{sblue}{HTML}{02BCD4}
\definecolor{sred}{HTML}{F44436}
\definecolor{spink}{HTML}{E91E62}
\definecolor{sgreen}{HTML}{8BC34A}
\definecolor{spurple}{HTML}{3F51B5}
\definecolor{slightgreen}{HTML}{CCDE3A}
\definecolor{sorange}{HTML}{FE9800}
\definecolor{sgolden}{HTML}{FFC108}

\usepackage{amsmath}
\usepackage{amssymb}
\usepackage{dsfont}
\usepackage{graphicx}
\usepackage{wrapfig,lipsum,booktabs}
\usepackage{xspace}
\usepackage{hyperref}

\begin{document}
\newcommand{\ours}{\texttt{FedImpres}\xspace}
\newcommand{\xiaoxiao}[1]{{\color{purple}[XL: #1]}}
\newcommand{\sana}[1]{{\color{blue}[SA: #1]}}

\title{Federated Impression for Learning with Distributed  Heterogeneous Data}


%
\titlerunning{3760}
%
\author{Atrin Arya\thanks{These authors contributed equally}\inst{1, 2}
\and
Sana Ayromlou$^*$\inst{1, 2}
\and
Armin Saadat\inst{1}
\and
Purang Abolmaesumi\inst{1}
\and
Xiaoxiao Li\inst{1, 2}
}
\authorrunning{Arya et al.}

\institute{
Electrical and Computer Engineering Department, The University of British Columbia, Vancouver, BC V6T 1Z4, Canada
    \and
    Vector Institute, Toronto, ON M5G 0C6, Canada 
    \email{\{atrinarya,s.ayromlou,xiaoxiao\}@ece.ubc.ca}}


%
%
%
%
\maketitle              
\begin{abstract}

Standard deep learning-based classification approaches may not always be practical in real-world clinical applications, as they require a centralized collection of all samples. Federated learning (FL) provides a paradigm that can learn from distributed datasets across clients without requiring them to share data, which can help mitigate privacy and data ownership issues.
In FL, sub-optimal convergence caused by data heterogeneity is common among data from different health centers due to the variety in data collection protocols and patient demographics across centers. Through experimentation in this study, we show that data heterogeneity leads to the phenomenon of catastrophic forgetting during local training.
We propose \ours which alleviates catastrophic forgetting by restoring synthetic data that represents the global information as federated impression. To achieve this, we distill the global model resulting from each communication round. Subsequently, we use the synthetic data alongside the local data to enhance the generalization of local training. Extensive experiments show that the proposed method achieves state-of-the-art performance on both the BloodMNIST and Retina datasets, which contain label imbalance and domain shift, with an improvement in classification accuracy of up to 20\%. The code is available at https://github.com/Atrin78/FedImpress.

\keywords{Federated Learning \and Catastrophic Forgetting \and Data Synthesis \and Data Heterogeneity.}
\end{abstract}
\section{Introduction}
\label{sec:intro}

Deep learning models are widely utilized in medical imaging owing to their promising outcomes. However, these models are typically designed for centralized environments where all data are stored in a single database.
Despite its benefits, centralizing data can be impractical for training purposes, \ie , healthcare facilities are generally hesitant to disclose their patients' information due to issues of data privacy, transmission costs, and access rights~\cite{o2004health}.
  Federated Learning (FL) presents a promising alternative, enabling multiple hospitals to leverage distributed data without sharing it. 
In each iteration, local models are initialized with the distributed server model. They are then trained on local data and send back their updates to the server for aggregation. However, conventional FL methods such as FedAvg
\cite{mcmahan2017communication} encounter performance degradation, when applied to non-IID (heterogeneous) data~\cite{karimireddy2020scaffold,li2019convergence}.

\begin{figure}[t]
	\centering
        	\includegraphics[width=0.8\textwidth]{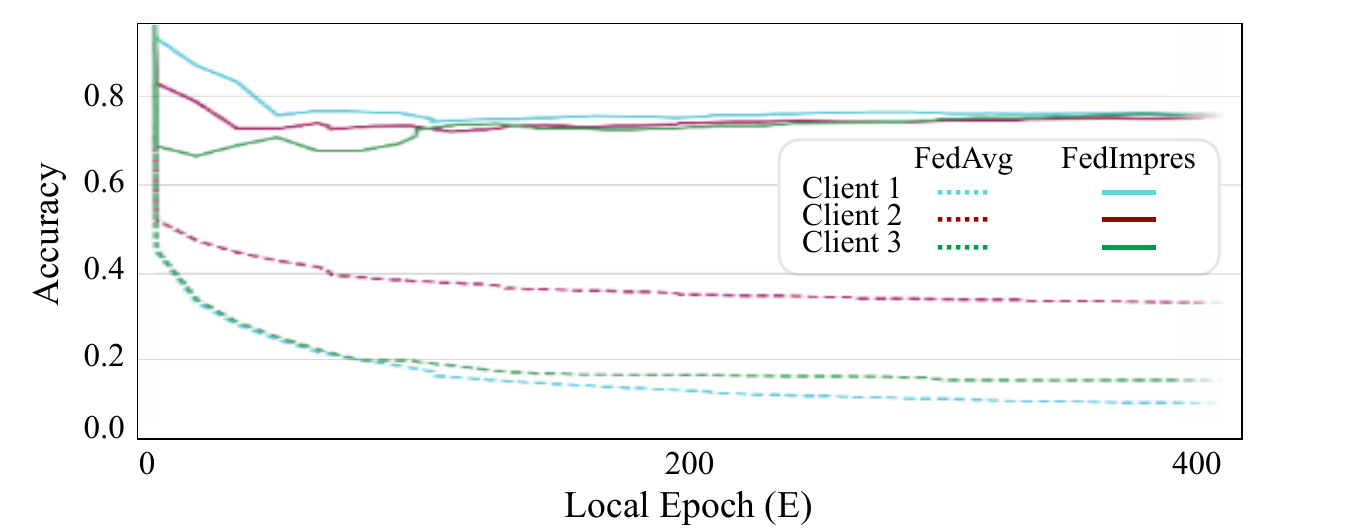}
        	\label{fig:CF}

	\caption{Catastrophic forgetting occurs when server weights are overwritten during local training, causing a loss of previous knowledge. To investigate the effect of catastrophic forgetting during local training in FL, we conducted experiments on BloodMNIST using the same experimental settings described in Sec.~\ref{sec:exp}. Specifically, we plot each client's local model accuracy over other clients' data during local training. The accuracy drops drastically using FedAvg; however, \ours maintains stable accuracy across clients.
     }
	\label{abstract}
\end{figure}

Heterogeneity happens due to 1) label imbalance \ie, various disease populations in different medical centers, and 2) domain shift, \ie, various data acquisition settings in medical devices. Studies have been carried out to mitigate each of the mentioned heterogeneities independently. However, based on our experiments in Fig.~\ref{abstract}, we show that both of these cases lead to a common issue called catastrophic forgetting, which has been usually overlooked in previous works. In FL, catastrophic forgetting~\cite{delange2021continual} occurs when a model overwrites past aggregated knowledge with local data. As shown in Fig.~\ref{abstract}, when observing a specific client during local training, the local model's accuracy on the other local datasets degrades since the server model's past aggregated knowledge is overwritten by the local heterogeneous data. In this work, we focus on solving the catastrophic forgetting issue in FL caused by label imbalance and domain shift.

Recent efforts in FL literature have mainly concentrated on improving local training on client side   \cite{jiang2022harmofl,li2020federated,liu2021feddg,zhu2022federated};  and refining aggregation on the server side \cite{li2021fedbn,wang2020federated,yeganeh2020inverse,luo2023gradma}. Notably, client side enhancements have been reported to achieve better outcomes~\cite{li2020federated}.
To improve client side training, two main categories of methodologies have been investigated: 1) \textit{model-level} approaches, which refine model optimization strategies through techniques such as setting a prior on model weights~\cite{li2020federated} or gradient update corrections~\cite{shoham2019overcoming,karimireddy2020scaffold}; and 2) \textit{data-level} methods which aim to alleviate statistical heterogeneity among local data across clients by employing techniques like sharing statistical information~\cite{dinsdale2022fedharmony,shin2020xor} or synthetic data generation~\cite{tang2022virtual,zhu2022federated}. Among them, model-level studies such as~\cite{shoham2019overcoming} and data-level studies such as~\cite{xu2022acceleration} have directly tackled the issue of catastrophic forgetting in FL. In terms of addressing catastrophic forgetting, data-level approaches exhibit superior model agnosticity, which is advantageous in deep learning~\cite{delange2021continual}. However, the generation of synthetic images with high fidelity that preserves the server model's information remains a persistent challenge.




In this paper, we propose a data-level approach, \ours, to mitigate catastrophic forgetting, caused by 
heterogeneous data in FL setting. To achieve this, after server aggregation in each FL iteration, we generate high-quality prototypical synthetic images by back-propagating on the server model's aggregated weights as a federated impression of global data. Furthermore, we add a model gradient-based constraint to this optimization to ensure that the synthesized data globally fits the entire latent distribution of the server model. 
 We share the synthesized data with clients and perform weighted training on both local and synthesized data on the client-side.
 We have chosen to use FedAvg as the base method for aggregating the local models on the server-side for the sake of simplicity. However, it is important to note that our approach is also compatible with other model aggregation strategies.




\section{Method}
\label{Methods}

\subsection{Problem setting}
The general FL setting aims to collaboratively train over a group of clients $\{C_1, C_2, ... ,C_N\}$ and their respective local datasets, with $N$ being the number of clients. The objective is to maintain high classification accuracy across all clients. Let $(x_i^n, y_i^n) \in \mathcal{X}_n$ represent an input image and its corresponding class label drawn from client $n$'s dataset. We denote the weights of feature extractors as $\theta$ and that of classifiers as $\phi$. In this setting, our goal is to have a model on the server that performs well on all clients by minimizing the following objective:

\begin{equation}
    J(\theta_G, \phi_G) =  \sum_{n=1}^{N} \mathbb{E}_{(x_i^n, y_i^n)\in \mathcal{X}_n}\ \ell(g(f(x_i^n;\theta_G); \phi_G), y_i^n),
\end{equation}
with loss function $\ell$ which is cross-entropy (CE) loss, $\mathcal{L}_{CE}$, in our case, client number $n$, server model's feature extractor $f(;\theta_G)$ and its classifier $g(;\phi_G)$. Note that the local data cannot be shared due to privacy concerns.
As a result, in each round $r$, we train models $\{f(;\theta^r_1), ... ,f(;\theta^r_N)\}$ initialized by $f(;\theta^r_G)$ using their respective client's local dataset, and share their weights $\{\theta^r_1, ..., \theta^r_N\}$ with the server model to aggregate them into $\theta^{r+1}_G$.
A common strategy for aggregation is~\cite{mcmahan2017communication} simply averaging the weights of clients, which we will follow in our study. 

\begin{figure}[t]
	\centering
	\includegraphics[width=0.9\textwidth]{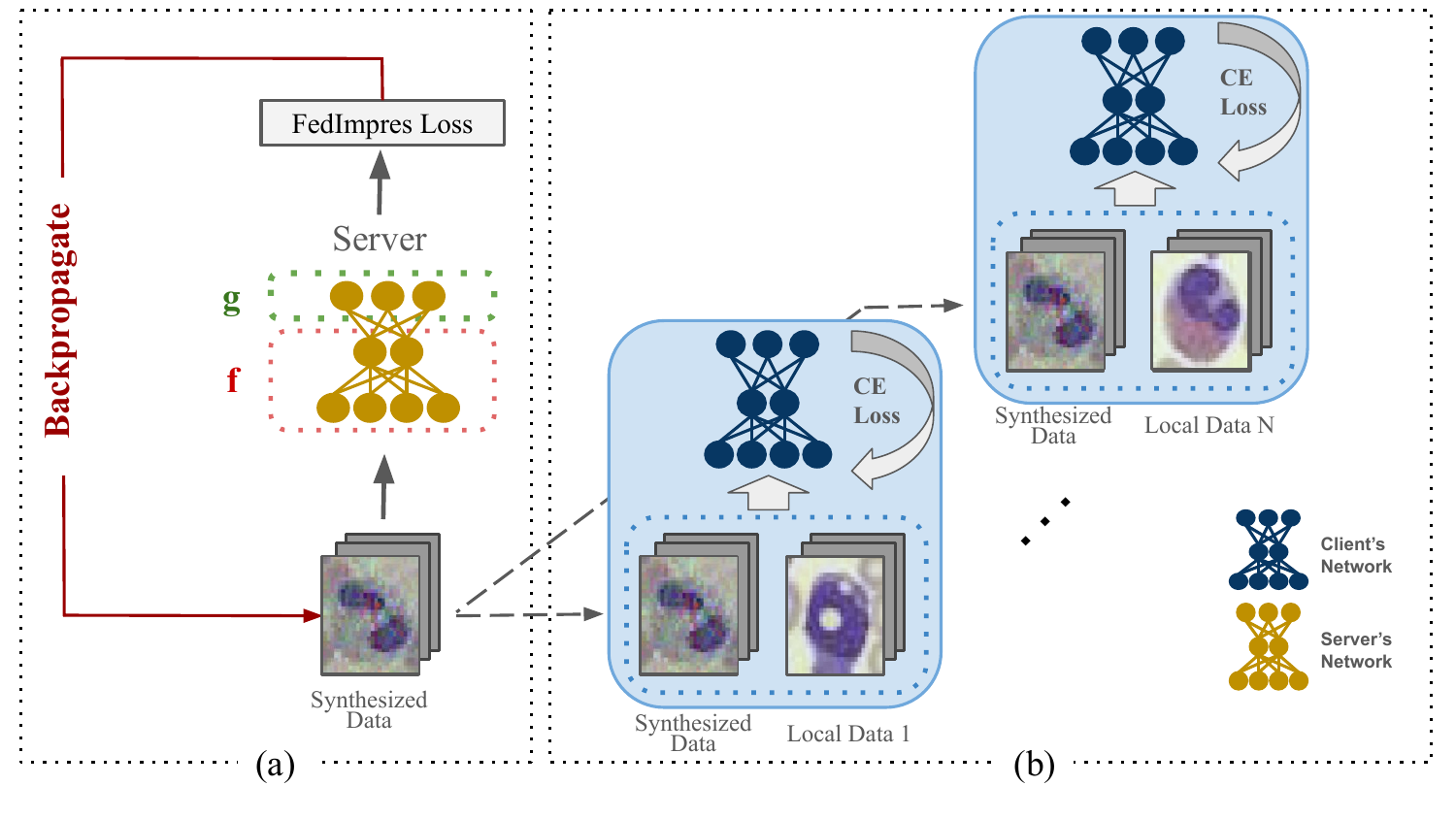}
	\caption{Our proposed approach, \ours, aims to capture the global distribution learned by the aggregated server model and distill it into a dataset that can be shared with clients. The approach consists of two steps: a) First, we perform pixel-wise optimization by starting from unlabeled public data and using the server model's predicted pseudo-labels to backpropagate using Eq.(\eqref{FI}).
    b) Second, to improve local training, we add the synthesized data as a regularizer to the local data using Eq.(\eqref{CE}). This allows us to share the global distilled distribution with clients and leverage it to improve local training.
 }
	\label{thresh}
\end{figure}

\subsection{Overview}
As described in the introduction (Sec.~\ref{sec:intro}), catastrophic forgetting during local training is one of the primary problems in heterogeneous FL. 
To develop a robust FL algorithm suitable for heterogeneous data, we need to address two fundamental challenges: 1) How to alleviate catastrophic forgetting in local training?  This can be achieved by utilizing a united synthetic data as a regularizer in local client training to penalize catastrophic forgetting; 2) How to generate this synthetic dataset? We can synthesize data using the server model to capture a genuine federated impression for local training. The overall paradigm of our method is shown in Fig.~\ref{thresh}. In the following sections, we will provide a detailed description of our proposed paradigm.


\subsection{Federated Impression}
Past methods like VHL~\cite{tang2022virtual} have proposed to use global synthetic data to improve FL on heterogeneous data. However, VHL's synthetic data does not preserve the server model's information useful for the targeted classification task during local training. Inspired by~\cite{ayromlou2022class}, to empower the global synthetic data to assist FL, we introduce an adaptive global data generation paradigm, which synthesizes data based on the server model in each communication round.
Next, we aim to have not only high-fidelity data but also the \textbf{information-preserving property}, \ie , training a model from scratch using synthesized data results in a model that performs similarly to the original server model. 
To obtain data with this characteristic, 
we optimize pixel values on the image space
$v_1, v_2, ..., v_S\in \mathcal{V}$ CE loss of the server model. Additionally, to achieve the information-preserving property following~\cite{yao2021source}, we add an equality constraint to the optimization process to ensure that the gradient of the server model's CE loss on $\mathcal{V}$ with respect to its weights $\theta^r_{G}$ is close to 0. Specifically, we aim to solve


\begin{equation}
\label{eq:constraint}
     \min_{\mathcal{V}}\ \sum_{(v_i, \hat{y}_i)\in\mathcal{V}} \mathcal{L}_{CE}(g(f(v_i;\theta^r_{G}); \phi^r_G), \hat{y}_i)\ s.t.\ \nabla_{(\theta^r_{G}, \phi^r_G)} \mathcal{L}_{CE} = 0 ,
\end{equation}
where $\hat{y}_i$ is initialized with the prediction of the server model when given $v_i$. Since optimizing Eq.~\ref{eq:constraint} is computationally expensive, according to~\cite{yao2021source}, we solve the relaxed version of the optimization problem imposing the equality constraint on $\phi^r_G$ only

\begin{equation}
\label{eq:constraint2}
     \min_{\mathcal{V}}\ \sum_{(v_i, \hat{y}_i)\in\mathcal{V}} \mathcal{L}_{CE}(g(f(v_i;\theta^r_{G}); \phi^r_G), \hat{y}_i)\ s.t.\ \nabla_{(\phi^r_G)} \mathcal{L}_{CE} = 0 ,
\end{equation}
It's worth noting that such a relaxation does not steer us away from our ultimate goal of information-preserving property. Instead of generating precise images with this property, we aim to produce images whose latent representation would capture the exact global distribution of the server in the latent space. Next, we solve it using the augmented lagrangian formulation:

\begin{center}
$\max_\Lambda \min_{\mathcal{V}}\ L_{ FedImpres} = $
\end{center}
\begin{equation}
    \sum_{(v_i, \hat{y}_i)\in\mathcal{V}} [\mathcal{L}_{CE}(g(f(v_i;\theta^r_{G});\phi^r_G), \hat{y}_i) + tr(\Lambda^T \nabla_{\phi^r_{G}} \mathcal{L}_{CE}) + \frac{\rho}{2}{||\nabla_{\phi^r_{G}} \mathcal{L}_{CE}||}^2],
    \label{FI}
\end{equation}
where $\Lambda$ is the lagrangian dual variable matrix 
for the equality constraint in Eq.~\eqref{eq:constraint} and $\rho$ is the penalty hyperparameter. According to~\cite{yao2021source}, we solve it approximately using an alternating direction method of multipliers (ADMM)~\cite{boyd2011distributed}. After synthesizing this data as the federated impression, we pass it to all clients for local training. Note that we don't need any additional private data information to generate the synthetic dataset compared to general FL methods like \cite{mcmahan2017communication}.

\subsection{Forgetting-Penalized Local Training}
To train the local model for client $n$, we receive an optimized synthetic dataset $\mathcal{V}$ from the server at the beginning of each local training round. To prevent catastrophic forgetting during local training, we train the model on synthetic data in addition to the local data using the following 
\begin{equation}
    \min_{(\theta^r_n, \phi^r_n)} \ L_{local}(\theta^r_n, \phi^r_n) + \beta L_{global}(\theta^r_n, \phi^r_n) ;
    \label{CE}
\end{equation}
where $L_{local}$ and $L_{global}$ are CE loss over each client's local data and shared global data, respectively. Here, $L_{global}$ basically used as a regularization term for improving the generalizability of local training over captured federated impression in the previous step.
This approach preserves information from the server model due to the information-preserving property of the synthetic data. Note that as opposed to~\cite{yao2021source}, we use the CE loss directly on the synthesized data to enforce the information-preserving property.
It is also worth noting that merely replacing the global loss with another regularization that instead aims to decrease the distance between the local model's and the server model's weights directly, as done in~\cite{shoham2019overcoming}, may not be optimal since it would limit the ability to capture local information.

\section{Experiments}
\label{sec:exp}

\subsection{Datasets}
We use two public medical image datasets to evaluate \ours{} on two typical heterogeneous settings for classification: label imbalance and domain shift:

\noindent\textbf{BloodMNIST} \cite{acevedo2020dataset} is one of the datasets in the standard medical imaging benchmark, MedMNIST \cite{medmnist}. We chose this dataset over other modalities as it contains adequate classes (eight in total), which can better demonstrate \ours on imbalanced labels settings. The images in this dataset are padded to size $32\times32$. 

\noindent\textbf{Retina} 
 dataset \cite{batista2020rim,diaz2019cnns,orlando2020refuge,sivaswamy2014drishti} consists of retina images of size $256\times 256$ gathered from four different sites, resulting in label imbalance and domain shift. We aim to solve the binary classification problem to detect Glaucomatous images from normal ones for this dataset. Samples and label distribution of both datasets for each client are provided in the supplementary material.

\subsection{Experimental Settings}
We conducted experiments to study label imbalance and domain shift 
among FL clients. For each experiment, we used three different alternatives of initialization for the synthesis step of \ours, \ie , random noise, public natural images (CIFAR-10~\cite{krizhevsky2009learning}), and a public \emph{unlabeled} medical dataset in a similar domain of local private data, which will be explained for each dataset separately. Note that obtaining unlabeled data from the same modality used for synthesis initialization is not a problem in the real world.\\
\noindent\textbf{Data Heterogeneity:} To simulate class imbalance, we used BloodMNIST. To replicate unlabeled medical data for synthesis initialization, we randomly selected 10\% of the data that were mutually exclusive from all of the training data. Afterwards, we utilized Latent Dirichlet Analysis (LDA)~\cite{he2020fedml,wang2020federated} to divide the remaining data into eight clients for an eight-way classification. We set the partition parameter of LDA ($\alpha$) to 0.01 and 0.005 to create moderate and severe imbalanced datasets. Subsequently, in a more practical evaluation, we carried out experiments on the Retina dataset, which encompasses data from four distinct domains with different demographic distributions and are naturally class-imbalanced. We employed data from one of the four sites as publicly accessible unlabeled data for synthesis initialization and performed binary classification on the remaining three datasets.\\
 \noindent\textbf{Implementation Details:} We used a simple Convolutional Neural Network (CNN) for classification in all settings. The architecture is detailed in the supplementary. All models were implemented with PyTorch and trained on
one NVIDIA Tesla V100 GPU with 16~GB of memory. 
Our implementation contains two stages of optimization in each communication round. 1) We freeze model weights for the image synthesis stage and use the SGD optimizer and optimize the batch of [16,32] images for 5 ADMM epochs in BloodMNIST and Retina, respectively. 2) In local model training, we update local model weights again with the SGD optimizer. We fixed the total training epochs for 400 iterations and performed our experiments in two different settings. We reported our results for 80 and 40 communication rounds with local update epochs (E) of 5 and 10, warmed up with 15 and 10 rounds of FedAvg, respectively. Hyperparameters are detailed in the Supplementary.

\begin{table*}[t]
\begin{center}
\begin{tabular}{l @{\hskip 0.1in} || c @{\hskip 0.1in} c@{\hskip 0.1in}  | c@{\hskip 0.1in}  c@{\hskip 0.1in}  | 
 c@{\hskip 0.1in}  c@{\hskip 0.1in}}
\hline
Dataset &\multicolumn{4}{c}{BloodMNIST} \vline  & \multicolumn{2}{c}{Retina}\\
\hline

 $\alpha$  & \multicolumn{2}{c}{0.01} \vline & \multicolumn{2}{c}{0.005} \vline & \multicolumn{2}{c}{NA}\\
\cline{1-7}
Local update epochs (E)  & 5 &  10  &  5 & 10  &  5 & 10  \\

\hline\hline

FedAvg \cite{mcmahan2017communication} & 83.1 & 82.4  & 39.0 & 37.6  & 55.7 & 52.0 \\
FedProx \cite{li2020federated} & 82.8 & 83.1  & 35.1 & 34.9  & 68.2 & 61.9 \\
VHL \cite{tang2022virtual} & \underline{84.9} & 83.3 & 50.3 & 43.0 & \underline{80.8} & 78.8 \\
FedVSS \cite{zhu2022federated} & 82.9 & 82.8  & 38.1 & 36.7  & 62.3 & 68.3 \\

FedCurv \cite{shoham2019overcoming} & 68.5 & 61.7  & 26.2 & 25.9  & 79.9 & 78.1 \\

FedReg \cite{xu2022acceleration} & 20.1 & 16.9  & 18.9 & 16.8  & 62.5 & 62.1 \\
\hline
\ours  (Random init) & 83.9 & 82.6  & 52.6 & 51.4  & 78.1 & \underline{80.6} \\
\ours  (CIFAR-10 init) & 84.1 & \underline{83.6}  & \underline{60.2 }& \underline{53.8}  & \textbf{81.5} & 79.8 \\

\ours (Medical init) & \textbf{94.2} & \textbf{93.1}  & \textbf{70.2} & \textbf{65.1}  & 80.6 & \textbf{81.1} \\
\hline
\end{tabular}
\end{center}
\caption{Classification accuracies on BloodMNIST and Retina dataset compared with the state-of-the-art methods. We reported \ours results using random, CIFAR-10 and medical unlabeled data of the same modality data as initialization. Although medical initialization performs overall better than CIFAR-10 and random, we still outperform baselines in most of the settings.}
\label{tab:results}
\end{table*}

\subsection{Comparison with Baselines}
We compared our results with common and state-of-the-art (SOTA) FL algorithms. Among common methods, we choose \textbf{FedAvg} \cite{mcmahan2017communication} and \textbf{FedProx} \cite{li2020federated} as two main baselines. FedProx solves performance degradation compared to FedAvg in the Non-IID setting by adding a regularization term for local training, which prevents divergence of local model weights from the server model. We also compare with SOTA FL methods that share similar ideas with ours by adding global synthetic data or editing local training. \textbf{VHL} \cite{tang2022virtual}, which generates global virtual data using untrained StyleGAN \cite{karras2019style} and does not update global virtual data during training. We also compare our results with \textbf{FedVSS} \cite{zhu2022federated}, which adversarially modifies local data using the server model to synthesize more general data for each client. Finally, we compare our results to SOTA methods \textbf{FedCurv} \cite{shoham2019overcoming} and \textbf{FedReg} \cite{xu2022acceleration} that focus on tackling the issue of catastrophic forgetting in FL.

The results are illustrated in Table~\ref{tab:results}. Although medical initialization has the best results, we show that even with CIFAR-10 and noise initialization, we outperform SOTA in most experiments, and this proves the effectiveness of the synthesis step regardless of the initialization. In all of the experiments \ours improves FedAvg by a large margin. This can be particularly observed when the level of heterogeneity is higher with $\alpha=0.005$ and the Retina dataset. Although FedProx was designed to have smoother local training by adding a penalty for divergence from the server model, this is harmful to severe heterogeneity due to a shortage in learning local data. Compared to VHL and FedVSS, we surpass them by virtue of our adaptive and unified synthesis data approach among clients, correspondingly. Although, FedCurv achieves close results to our method on Retina dataset, its performance degrades when facing label shift on the BloodMnist dataset. FedReg does not perform well on both datasets since it's not designed for architectures with batch normalization.

\subsection{Ablation Studies}


\begin{table*}[t]
\caption{Classification accuracies reported on the Retina dataset comparing synthesizing with \ours (CE loss) (Eq.~\ref{eq:constraint}) vs. vanilla CE loss. In both cases, we initialize the synthesis step with random noise.}\label{tab:Ablation}

\begin{tabular}{l @{\hskip 0.1in} || 
 c@{\hskip 0.1in}  c@{\hskip 0.1in} }
\hline
Dataset  & \multicolumn{2}{c}{Retina}\\

\cline{1-3}
Local update epochs (E) &  5 & 10   \\

\hline\hline
Data synthesis w CE loss  & 73.9 & 75.4 \\
Data synthesis w FedImpres loss &  \textbf{78.1} & \textbf{80.6} \\


\hline

\end{tabular}

\end{table*}

To assess the effect of our data synthesis algorithm, we consider another synthetic data generation variant adopted by our proposed method and study its performance on the Retina dataset, as it is a real-life dataset and has both label imbalance and domain shift.
For this, we omit the constraint of globalizing data synthesized to distribution seized by the server model in Eq.~(\ref{eq:constraint}) and optimize only with CE loss. For both methods, we use random noise to initialize data synthesis to omit any initialization bias. As shown in Table~\ref{tab:Ablation}, the proposed \ours approach surpasses its other variant, showing the effectiveness of its data synthesis algorithm for data generation.


\section{Conclusion}
Previous FL approaches suffer from catastrophic forgetting in their local training due to the heterogeneity of the distributed data. This problem becomes more pronounced for clients dealing with medical data due to the heterogeneity caused by both domain shift and label imbalance across clients. To this end, we proposed a novel method called \ours, which uses the server model to generate synthetic data at each round to account for the server model's information in the local training and avoid forgetting. We demonstrated how this method could achieve superior performance for two benchmark medical datasets, particularly in highly heterogeneous cases. Moreover, the ablation section showed the data synthesis algorithm's effectiveness. It is worth noting the synthetic data-restoring method is efficient without training additional generative models.
Furthermore, our proposed method shows the potential to be applied in many healthcare applications using data from multiple centers. We will explore integrating our research with other practical applications in the medical domain. This may involve testing our approach on various medical datasets and improving the pipeline to meet the preferred standards of clinical practice. 
\begin{credits}
\subsubsection{\ackname}
This work is supported in part through computational resources and services provided by Advanced Research Computing at the University of British Columbia, Natural Sciences and Engineering Research Council of Canada (NSERC), the Canadian Institutes of Health Research
(CIHR), Compute Canada, and Vector Institute.
\subsubsection{\discintname}
The authors have no competing interests to declare that are relevant to the content of this article.
\end{credits}


%
%
%
\bibliographystyle{splncs04}
\bibliography{reference.bib}

\newpage
\appendix
\subsection*{Dataset Visualization \& Details}
\begin{figure}[H]
\centering
\centering
         \begin{subfigure}[t]{0.7\textwidth}
\includegraphics[width= 1 \textwidth]{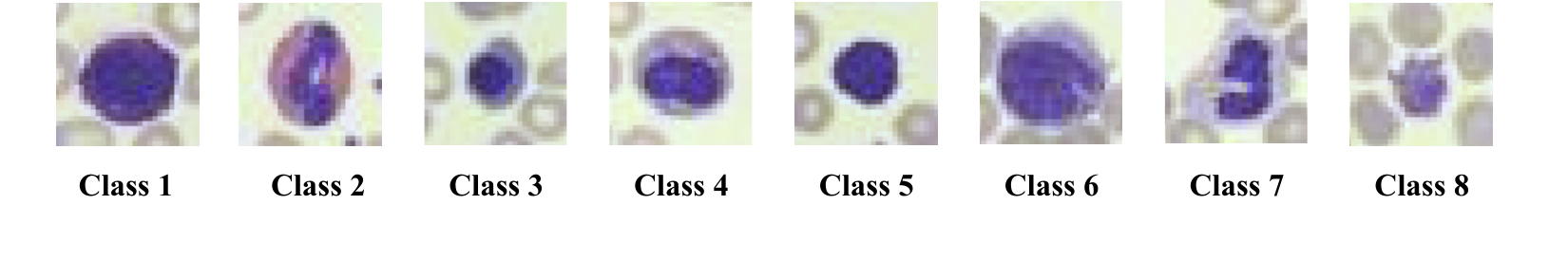}
\end{subfigure}
         \begin{subfigure}[t]{0.7\textwidth}
\includegraphics[width= 1 \textwidth]{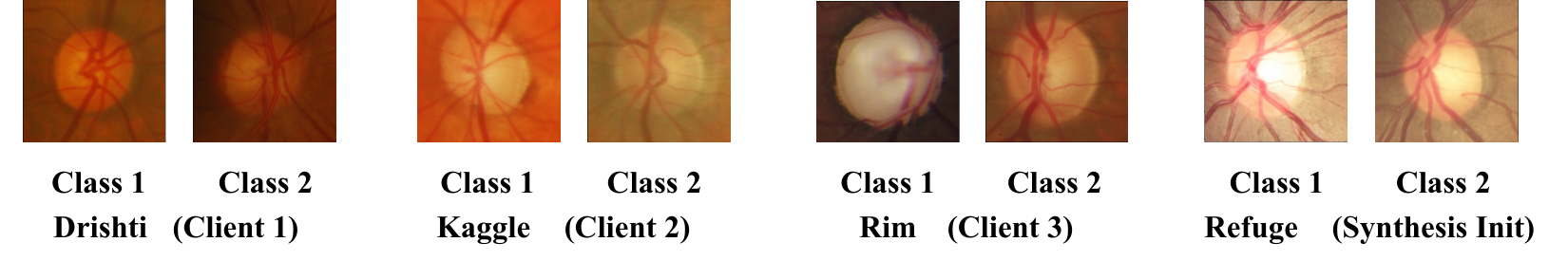}
\end{subfigure}
\caption{Top: BloodMNIST, consisting of eight classes; bottom: Normal (class 1) and Glaucomatous (class 2) images from the Retina, collected from each site.}
\end{figure}
\vspace{-1cm}

\begin{table}[H]
\centering
\resizebox{0.96\columnwidth}{!}{
\begin{tabular}{l || c c c c c c c c | c c c c c c c c | c c }
\hline
Dataset &\multicolumn{16}{c}{BloodMNIST} \vline&\multicolumn{2}{c}{Retina}\\
\hline

 $\alpha$  & \multicolumn{8}{c}{0.01} \vline & \multicolumn{8}{c}{0.005}  \vline & \multicolumn{2}{c}{NA}\\
\hline
Classes  & 1& 2& 3& 4& 5& 6& 7& 8& 1& 2& 3& 4& 5& 6& 7& 8 & 1 & 2 \\

\hline

Client 1  & 5 & 0 & 0 & 0 & 22 & 577 & 185 & 58 & 11 & 0 & 0 & 0 & 0 & 0 & 0 & 0 & 61& 21  \\
Client 2  &  0 & 386 & 67 & 2 & 73 & 304 & 3 & 0 & 0 & 0 & 533 & 0 & 0 & 0 & 0 & 0 & 49& 33 \\
Client 3  & 0 & 0 & 0 & 0 & 0 & 0 & 1195 & 181 & 0 & 0 & 0 & 996 & 0 & 0 & 0 & 0  & 30& 52 \\
Client 4  & 0 & 120 & 0 & 0 & 6 & 1 & 0 & 768& 0 & 0 & 0 & 0 & 0 & 483 & 0 & 0 & \multicolumn{2}{c}{NA} \\
Client 5  & 0 & 28 & 6 & 0 & 4 & 1 & 687 & 486& 0 & 0 & 0 & 0 & 0 & 0 & 1207 & 0 & \multicolumn{2}{c}{NA}  \\
Client 6  & 0 & 1 & 705 & 0 & 661 & 0 & 0 & 0& 407 & 0 & 0 & 0 432 & 0 & 0 & 0 & 0 & \multicolumn{2}{c}{NA}   \\
Client 7  & 760 & 1418 & 0 & 0 & 0 & 0 & 0 & 0& 0 & 0 & 0 & 0 & 1 & 0 & 0 & 856 & \multicolumn{2}{c}{NA}  \\
Client 8  & 1 & 8 & 208 & 1836 & 0 & 0 & 1 & 0& 0 & 1054 & 0 & 0 & 0 & 0 & 0 & 0 & \multicolumn{2}{c}{NA}  \\
\hline
\end{tabular}
}
\caption{Number of data points of each class for each client.}
\label{tab:distribution}
\end{table}

	

\subsection*{Model Architecture \& Hyperparameters Details}
\begin{table}[H]
    \begin{subtable}[h]{0.25\textwidth}
    \begin{center}
        \begin{tabular}{c|c}
\toprule
\textbf{L}              & \textbf{Details}    \\ \midrule  
\multirow{2}{*}{1}& Conv2D(3, 64, 5, 1, 2)                    \\ 
& BN(64), ReLU, MaxPool2D(2,2) \\ \hline
\multirow{2}{*}{2}& Conv2D(64, 64, 5, 1, 2)                    \\ 
& BN(64), ReLU, MaxPool2D(2,2) \\ \hline
\multirow{2}{*}{3}& Conv2D(64, 128, 5, 1, 2)                    \\ 
& BN(64), ReLU\\ \hline
4& FC(8192,2048), BN(2048), ReLU\\ \hline
5& FC(2048,512), ReLU\\ \hline
6& FC(512,8)                \\ \hline
\end{tabular}
\end{center}
    \end{subtable}
     \hfill
     \hfill
    \begin{subtable}[h]{0.55\textwidth}
    \begin{center}
        \begin{tabular}{c|c}
\toprule
\textbf{L}              & \textbf{Details}    \\ \midrule  
1   & Conv2D(3, 64, 11, 4, 2)                    \\ 
& BN(64), ReLU, MaxPool2D(3,2)   \\ \hline
2 & Conv2D(64, 192, 5, 1, 2)                    \\ 
& BN(192), ReLU, MaxPool2D(3,2)                   \\ \hline
3 & Conv2D(192, 384, 3, 1, 1),                     BN(384), ReLU                   \\ \hline
4 &   Conv2D(384, 256, 3, 1, 1),                     BN(256), ReLU                     \\ \hline
5  & Conv2D(256, 256, 3, 1, 1), BN(256) \\ & ReLU, MaxPool2D(3,2), AvgPool2D \\ \hline
6 &   FC(9216,4096), BN(4096), ReLU                 \\ \hline
7 & FC(4096,4096), BN(4096), ReLU                  \\ \hline
8 & FC(4096,2)         \\ \bottomrule  
\end{tabular}
\end{center}
     \end{subtable}
\caption{
The architecture of the benchmark experiment includes specific parameters for each layer type. For Conv2D layers, we list the input and output dimensions, kernel size, stride, and padding. For MaxPool2D layers, we list the kernel size and stride. For FC layers, we list the input and output dimensions. For BN layers, we list the channel dimension. Right and left architecture is used for BloodMNIST and Retina, respectively.
}
\label{tab:arch}
\end{table}

\label{app_data}

\begin{table}
\centering
\begin{tabular}{c|c| c c | c c | c c}

\hline
\multirow{3}{*}{\textbf{Stage}} & Dataset &\multicolumn{4}{c}{BloodMNIST} \vline &\multicolumn{2}{c}{Retina}\\
\cline{2-8}

&Local Update Epochs(E) & 5& 10& 5& 10 & 5 & 10 \\

\hline
\hline

\multirow{2}{*}{\textbf{Image Synthesis}} & Learning rate  & 0.1 & 0.1 & 0.1 & 0.1 & 0.01 & 0.01\\ \cline{2-8}
& Batch size & 16 & 16 & 16 & 16 & 20 & 40\\ \cline{2-8}
\hline
\multirow{3}{*}{\textbf{ADMM}} & Iteration numbers & 5 & 5 & 5 & 5 & 5 & 5\\ \cline{2-8}
& $\rho$ & 0.2 & 0.2 & 0.2 & 0.2 & 0.2 & 0.2\\ \cline{2-8}
& Gamma param & 0.01 & 0.01 & 0.01 & 0.01 & 0.001 & 0.001\\ \cline{2-8}
\hline
\multirow{3}{*}{\textbf{Training}} & Learning rate  & 0.01 & 0.01 & 0.01 & 0.01 & 0.01 & 0.01\\ \cline{2-8}
 & Batch size & 16 & 16 & 16 & 16 & 30 & 60\\ \cline{2-8}
  & $\beta$ ($L_{reg}$ coefficient) & 1 & 1 & 1 & 1 & 0.5 & 0.5\\ \cline{2-8}
\hline

\end{tabular}
\caption{ Hyper-parameters used for training.}
\label{tab:hyp}
\end{table}

\end{document}